\pdfoutput=1

\documentclass[11pt]{article}

\usepackage[]{acl}

\usepackage{times}
\usepackage{latexsym}
\usepackage{amsmath}  
\usepackage{amssymb}  
\usepackage{multirow}
\usepackage{graphicx}
\usepackage{adjustbox}
\usepackage{booktabs} 
\usepackage{float}
\usepackage{caption}
\usepackage{subfigure} 
\graphicspath{{./figures/}}

\usepackage[T1]{fontenc}

\usepackage[utf8]{inputenc}

\usepackage{microtype}

\usepackage{inconsolata}
\title{Qilin-Med-VL: Towards Chinese Large Vision-Language Model for General Healthcare}

\author{%
   Junling Liu$^{\dagger}$\thanks{ \ \  Corresponding Author. $^{\dagger}$Co-first authors}  \
  Ziming Wang$^{\dagger}$
  Qichen Ye$^{1\dagger}$ 
  Dading Chong$^{1\dagger}$ 
  Peilin Zhou$^{2\dagger}$ 
  Yining Hua$^{3\dagger}$ 
  \\
\\
  \texttt{william.liuj@gmail.com, wang.zm@pku.edu.cn, yeeeqichen@pku.edu.cn} \\
  \texttt{1601213984@pku.edu.cn, zhoupalin@gmail.com, yininghua@g.harvard.edu}
}

\begin{document}
\maketitle
\begin{abstract}
Large Language Models (LLMs) have introduced a new era of proficiency in comprehending complex healthcare and biomedical topics.  However, there is a noticeable lack of models in languages other than English and models that can interpret multi-modal input, which is crucial for global healthcare accessibility. In response, this study introduces \textbf{Qilin-Med-VL}\footnote{Materirals of this study are available at \url{https://github.com/williamliujl/Qilin-Med-VL}} 
, the first Chinese large vision-language model designed to integrate the analysis of textual and visual data.  
Qilin-Med-VL combines a pre-trained Vision Transformer (ViT) with a foundational LLM. It undergoes a thorough two-stage curriculum training process that includes feature alignment and instruction tuning.  This method enhances the model's ability to generate medical captions and answer complex medical queries.
We also release \textit{ChiMed-VL}, a dataset consisting of more than 1M image-text pairs. This dataset has been carefully curated to enable detailed and comprehensive interpretation of medical data using various types of images.

\end{abstract}

\section{Introduction}
The introduction of Large Language Models (LLMs) into the field of healthcare and biomedicine has brought significant advancements. GPT-4's notable achievement on the United States Medical Licensing Examination (USMLE) demonstrates its proficiency in complex biomedical concepts and its potential as a tool for healthcare professionals \cite{nori2023capabilities}. This milestone reflects the model's extensive training and knowledge, as well as the potential of medical LLMs.

However, the practical implementation and support for making decisions in healthcare and biomedicine require the use of multi-modal techniques due to the intricate nature of medical diagnosis and patient care \cite{Yu2018ArtificialII}. Determinative information often goes beyond written content to encompass visual indicators of how illnesses manifest. Disease patterns, clinical diagnoses, and many other aspects often rely on the analysis of visual cues. This includes patterns of skin lesions for dermatological conditions \cite{wan2022,skin2023} or the interpretation of electrocardiograms and brain scans for cardiac and neurological issues \cite{Yu2018ArtificialII, brain}. Chronic conditions like diabetes require analysis of visual data like retinal scans \cite{Gupta2022RealTA}, while cancer treatment benefits from detailed imaging from CT scans or MRIs\cite{khoo1997magnetic}. This highlights the limitations of relying solely on textual data and emphasizes the demand for integrated methods that combine visual data analysis with human-like conversation.

In the English world, \cite{Li2023LLaVAMedTA} has undertaken a pioneering endeavor to develop LLaVA-Med, an LLM that combines advanced visual-textual data analysis in the field of biomedicine through a process of multistaged multi-modal instruction tailoring. However, it is crucial to recognize that language barriers persist as a significant impediment to the advancement of global health \cite{linguisticsbarrierhealthcare}. A shortsighted focus on English-centric models could exacerbate inequalities in healthcare accessibility. 

Given the current absence of large vision-language models for Chinese medical fields, we reduce this inequality by working on developing Chinese healthcare and biomedical models, recognizing the significant impact that linguistic inclusion has on improving global health standards. Expanding on this foundation, our research introduces pivotal contributions:
\begin{enumerate}
    \item \textbf{Qilin-Med-VL}, the first large Chinese medical vision-language model, proficient in multiple critical medical imaging disciplines.
    \item The first large-scale Chinese Vision-Language dataset for general healthcare, \textbf{Chi}nese \textbf{Med}icine - \textbf{V}ision \textbf{L}anguage \textit{ChiMed-VL}, designed to facilitate multistage training. This dataset has two subsets: vision-language feature alignment and instruction tuning.
\end{enumerate}
Models like Qilin-Med-VL look forward to helping healthcare professionals make better decisions by providing them with more insights. Ultimately, our goal is to improve healthcare worldwide. We believe that our work represents a new frontier in research, where technology and medical knowledge come together to create a brighter and more equitable future for healthcare.
\begin{figure*}[ht]
\centering
\includegraphics[width=0.8\textwidth]{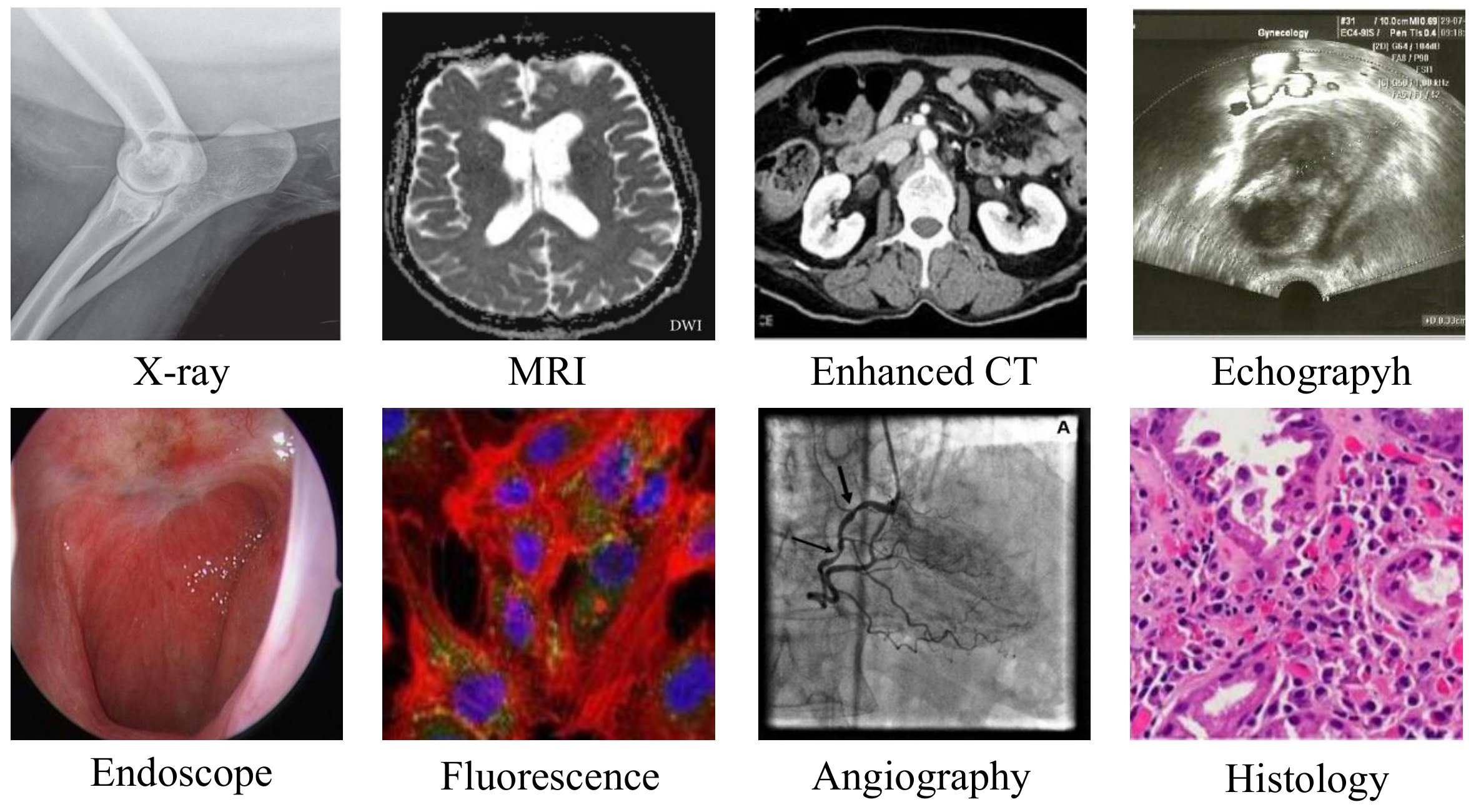}
\caption{Samples of various types of medical images in ChiMed-VL dataset.}
\label{fig:image_samples}
\end{figure*}

\begin{table*}[htb]
\caption{Basic statistics of ChiMed-VL-Alignment. C: Contexts; I: Inlines.}
\centering
\resizebox{0.7\textwidth}{!}{
\begin{tabular}{lcccc}
\hline
 & {PMC-CaseReport} & {PMC-OA} & {Total} \\
\hline
Image-Text pairs  \#& 316,838 & 263,176 & 580,014 \\
C Tokens  \# & 167M & - & 167M \\
I Tokens  \# & 21M & 42M & 63M \\
Max C tokens & 2,576 & - & 2,576\\
Max I tokens & 1,551 & 1,417 & 1,551  \\
Median (Q1, Q3) C tokens & 435 (211, 757) & - & 435 (211, 757)  \\
Median (Q1, Q3) I tokens & 59 (41, 83) & 125 (68, 210) & 75 (47, 132)  \\
\hline
\end{tabular}}
\label{tab:basic_stats_of_ChiMed-VL-Alignment}
\end{table*}

\begin{figure*}[htbp]	
	\subfigure[] 
	{
		\begin{minipage}{\columnwidth}
			\centering          
			\includegraphics[width=0.75\textwidth]{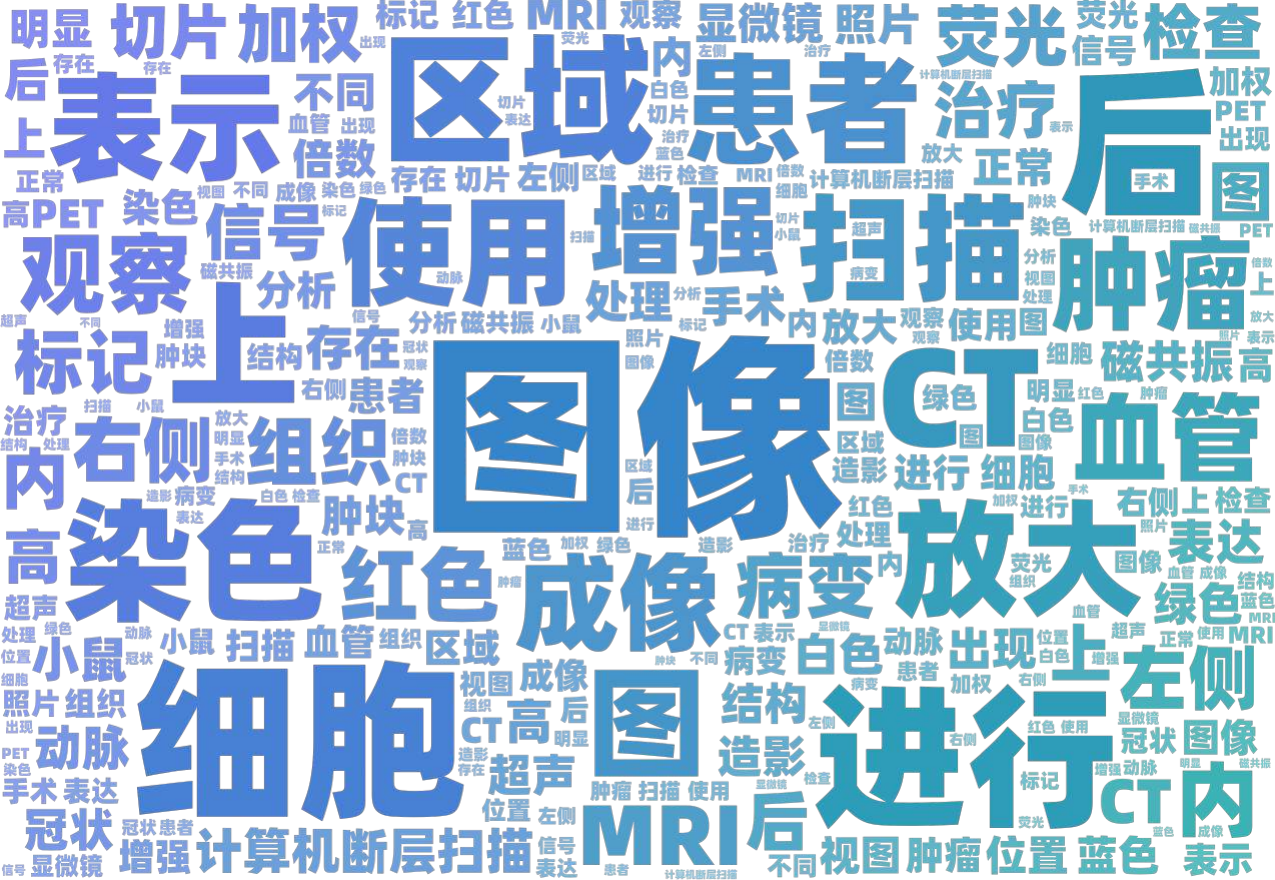}   
		\end{minipage}
	}
	\subfigure[] 
	{
		\begin{minipage}{\columnwidth}
			\centering      
			\includegraphics[width=0.75\textwidth]{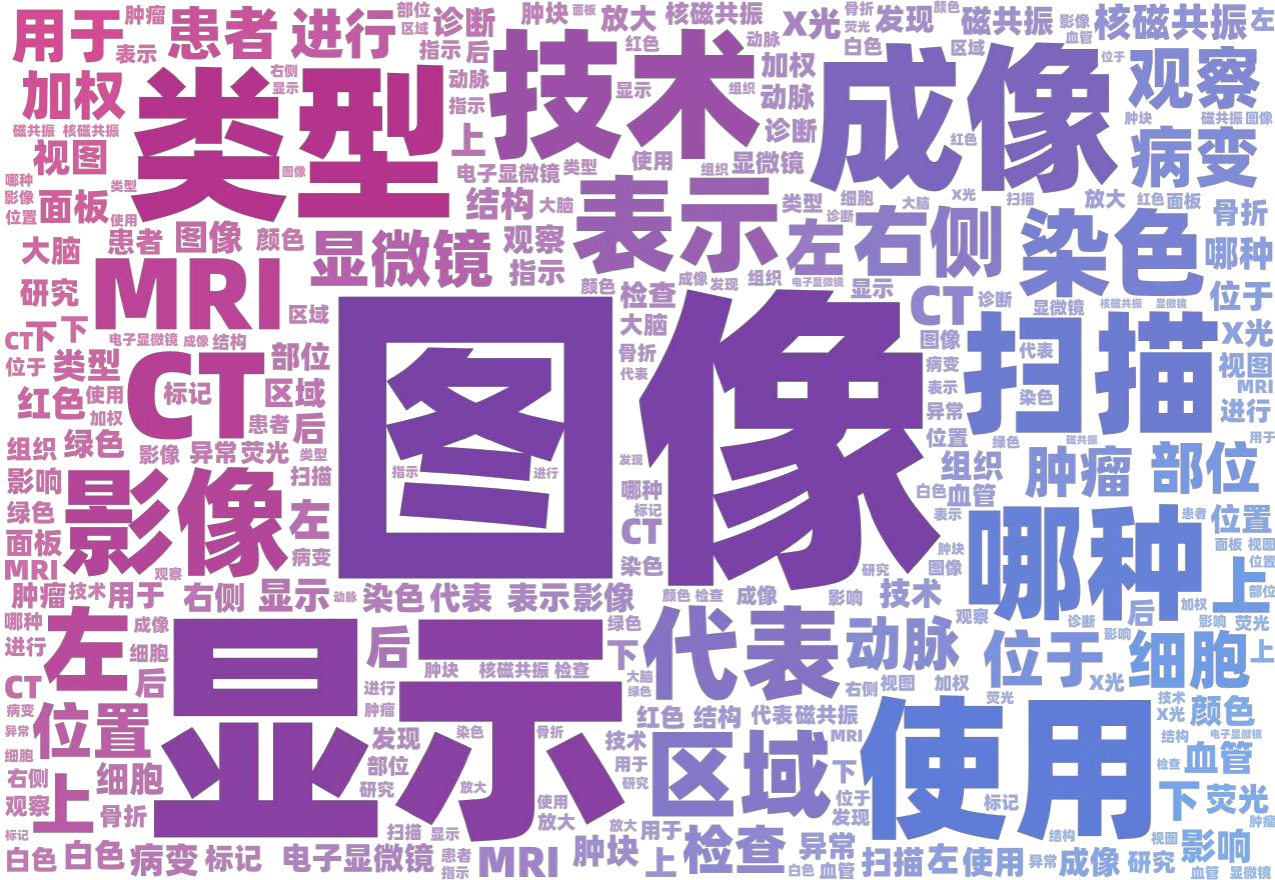}   
		\end{minipage}
	}
	\caption{(a) Word cloud of ChiMed-Alignment. (b) Word cloud of ChiMed-Instruction.} 
	\label{fig:word-cloud}  
\end{figure*}
\begin{table*}[htb]
\caption{Basic statistics of ChiMed-VL-Instruction. Q: Questions; A: Answers.}
\centering
\resizebox{0.7\textwidth}{!}{
\begin{tabular}{lcccc}
\hline
 & {PMC-CaseReport} & {PMC-VQA}  & {Total} \\
\hline
QA pairs \# & 316,838 & 152,603 & 469,441 \\
Q Tokens  \# & 7M & 3M & 10M \\
A Tokens \# & 10M & 3M & 13M \\
Max Q tokens & 4,040 & 335 & 4,040  \\
Max A tokens & 451 & 732 & 732  \\
Median (Q1, Q3) Q tokens & 21 (16, 26) & 18 (15, 22) & 20 (16, 25)  \\
Median (Q1, Q3) A tokens & 27 (20, 38) & 10 (6, 16) & 22 (12, 34)  \\
\hline
\end{tabular}}
\label{tab:basic_stats_of_ChiMed-VL-Instruction}
\end{table*}

\section{Related Work}

\subsection{Multi-modal LLMs}
The advent of LLMs has transformed the field of multi-modal LLMs field, which now has a branch that focuses on the adaptability of LLMs to incorporate various modalities. 
For example, AnyMal\cite{Moon2023AnyMALAE} generates textual responses from input signals, including text, image, video, audio, and IMU motion sensor. NExT-GPT\cite{Wu2023NExTGPTAM} accomplishes universal multi-modal comprehension and diverse input/output modalities by integrating LLM with multi-modal adaptors and diffusion decoders. A typical focus of this field is on integrating visual elements, which is primarily concerned with integrating vision as a `foreign language' into the models, and can thus be easily adapted to other modalities. These models are typically referred to as large vision-language models.

Pioneering research, such as Flamingo \cite{Alayrac2022FlamingoAV}, highlights the effectiveness of these models in synthesizing visual and textual information, resulting in nuanced, unrestricted content. Noteworthy developments like the Q-Former by BLIP-2 \cite{Li2023BLIP2BL} contribute to harmonizing pre-trained vision models with LLMs, driving forward the capabilities of these systems.

Models like MiniGPT-4 \cite{Zhu2023MiniGPT4EV} and LLaVA \cite{li2023llava} laveraged GPT-4 create conversational visual instruction datasets. These datasets enhance the models' proficiency in correlating visual traits with linguistic elements. LLaVA-1.5 \cite{Liu2023ImprovedBW} has advanced through strategic enhancements and high-performance standards in multi-modal LLM evaluations. It outperformed many open-source models, demonstrating significant improvements.  


Meanwhile, VisCPM \cite{Hu2023LargeMM}, InternLM-XComposer \cite{Zhang2023InternLMXComposerAV}, and Qwen-VL \cite{Bai2023QwenVLAF}, have excelled in interpreting and executing instructions in Chinese, reflecting the global applicability and adaptability of these advanced systems. These achievements not only showcase the models' versatility in processing language-specific tasks but also highlight their capability to handle intricate instructions across various domains and applications.



\subsection{Large Medical Vision-Language Models}

Research in large medical vision-language models has been encouraging, with significant efforts put into establishing foundational models. Noteworthy initiatives include LLaVA-Med \cite{Li2023LLaVAMedTA} and MedVInT \cite{Zhang2023PMCVQAVI}, which utilize image captions from PubMed Central \cite{doi:10.1073/pnas.98.2.381} for fine-tuning medical visual instruction sets. 

Medical visual question answering (VQA) has received extensive attention and research due to its substantial practical uses. Pushing the boundaries of medical VQA capabilities, Med-Flamingo \cite{Moor2023MedFlamingoAM} emerged with capabilities for few-shot generative medical VQA on interleaved medical image-text data.  Additionally, MedBLIP \cite{Chen2023MedBLIPBL} narrows its focus to a specialized segment of 3D imaging, primarily MRI.   

Beyond medical VQA, Med-PaLM M \cite{Tu2023TowardsGB}, adopted an innovative approach, proposed a generalist biomedical AI system that can perform medical image classification, medical VQA, radiology report generation and summarization, and more with the same set of model weights. In radiological diagnostics, models like RadFM \cite{Wu2023TowardsGF} and Radiology-Llama2 \cite{Liu2023RadiologyLlama2BL} demonstrated promising performance in enhancing diagnostic precision through visual aid.

Despite these advances, a research gap persists concerning Chinese medical LLMs tailored for multi-modal inputs. Existing models, such as Huatuo \cite{Wang2023HuaTuoTL}, Qilin-Med \cite{Ye2023QilinMedMK}, and CMExam \cite{Liu2023BenchmarkingLL} only allow textual inputs.  Bridging this gap is crucial, considering the potential impact on healthcare accessibility, where linguistic barriers can restrict critical information and services. This concern is especially pronounced for non-mainstream language speakers currently underserved by prevalent NLP technologies \cite{bird2020decolonising, zeng2022greenplm}. Prioritizing such research is imperative to mitigate systemic disparities and democratize access to crucial healthcare advancements.

 In this work, we harness these advancements to develop a specialized Chinese medical vision-language model, characterized by its efficiency and efficacy in operation.

\section{Dataset Construction}





Addressing the scarcity of Chinese medical multi-modal data for training instruction-following models, we introduce a pioneering dataset, \textit{ChiMed-VL}. \textit{ChiMed-VL} was established by leveraging several open-source English medical multi-modal datasets. We translated these datasets into Chinese with GPT-3.5 and conducted expert quality control. The dataset contains two components: concept alignment and instruction-following, each critical during distinct training phases.

\subsection{The Concept Alignment Subset}


To enable model support for a multitude of medical image types, we leveraged two comprehensive open-source multi-modal medical datasets: \textit{PMC-OA} \cite{lin2023pmc}  and \textit{PMC-CaseReport} \cite{PMC-CaseReport}. These datasets collectively cover an extensive range of diagnostic modalities, such as X-ray, MRI, CT, Radioisotope, Mitotic, and several others, examples of which are depicted in Fig.\ref{fig:image_samples}. Recognizing the disparity induced by the scarcity of Chinese-centric data, we used GPT-3.5 to translate the dataset into Chinese. The breakdown of this translation process is elaborated in Tab.\ref{tab:basic_stats_of_ChiMed-VL-Alignment} and Fig.\ref{fig:word-cloud}(a).

\textit{ChiMed-VL-Alignment} consists of 580,014 image-text couplings, each pair falling into one of two categories: context information of an image or descriptions of an image. The context category contains 167M tokens, presenting a median text length of 435 (Q1: 211, Q3: 757). Conversely, descriptions, more concise and image-specific, contain inline descriptions and captions. They comprise 63M tokens, with median lengths settling at 59 (Q1: 45, Q3: 83).


\subsection{The Instruction-Tuning Subset}


In the second phase, we constructed the \textit{ChiMed-VL-Instruction} subset for refining the model's interpreting and instruction following capabilities. We extracted data from two open-source compilations: \textit{PMC-Report} and \textit{PMC-VQA} \cite{zhang2023pmc}. These datasets contain a diverse collection of data, including X-rays, CT scans, Echography, and Ultrasonography, enriching the model's familiarity with varied medical scenarios. We again used GPT-3.5 to translate the English questions and their corresponding answers into Chinese. Tab.\ref{tab:basic_stats_of_ChiMed-VL-Instruction} and Fig.\ref{fig:word-cloud}(b) details the statistics of this process.

\textit{ChiMed-VL-Instruction} comprises 469,441 question-answer pairs. Within this subset, the questions section contains 10M tokens with a median length of 20 (Q1: 16, Q3: 25), posing a concise inquiry reflective of medical queries. The answers consist of 13M tokens with a median length slightly longer at 22 (Q1: 12, Q3: 34), providing clear, direct, and informative responses.

\begin{figure*}[htb]
\centering
\includegraphics[width=\textwidth]{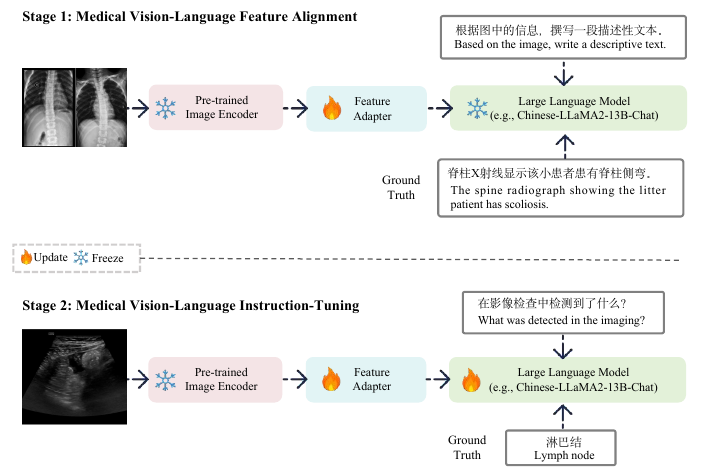}
\caption{The two-stage curriculum training scheme of Qilin-Med-VL.}
\label{fig: flowchart}
\end{figure*}


\subsection{Data Pre-processing}

A significant challenge addressed during the compilation of ChiMed-VL involved the management of input images. Datasets from \textit{PMC-OA} and \textit{PMC-CaseReport} contain multiple images corresponding to single text snippets. To enhance medical visual concept alignment and mitigate potential misalignments, images related to the same text were concatenated into single composite images, forming unified image-text pairs. This method necessitated the exclusion of samples with more than four images per text to avoid low-resolution outputs post-concatenation. We strategically chose horizontal or vertical image concatenation based on the original image sets' dimensions, preventing extreme aspect ratios in the combined image. Furthermore, we discarded samples with overly brief textual descriptions or those impractical for translation. 

The final training data format emulates a conversation between an assistant and an individual providing visual instructions, structured via task-specific Chinese prompts. Approximately 20 unique prompt templates were designed for each task, ensuring a diverse training schema. For each sample, a template was randomly selected from the task-specific set, and the data was reformulated into a dialog format, making it a robust resource for training purposes.

\label{stage2_data}

\section{Methodology}
\subsection{Overall Architecture}

The Qilin-Med-VL architecture comprises three key components:
\begin{enumerate}
    \item \textbf{Foundation LLM}: Qilin-Med-VL employs the renowned Chinese LLM, Chinese-LLaMA2-13B-Chat, to comprehend linguistic content and generate appropriate responses.
    \item \textbf{Pretrained Image Encoder}: To process medical images, Qilin-Med-VL leverages the Vision Transformer (ViT)~\cite{dosovitskiy2021image} to obtain visual features from each image.
    \item \textbf{Vision-Language Feature Adapter}: This component aims to align visual features with linguistic features, creating a shared feature space to effectively capture complementary information from different modalities. For efficiency, a simple linear projection layer is used as the feature adapter. In the future, we plan to investigate more effective and sophisticated adapters.
\end{enumerate}

\subsection{Two-stage Curriculum Training Scheme}
As shown in Fig.~\ref{fig: flowchart}, the training procedure of Qilin-Med-VL is divided into two stages: vision-language feature alignment and instruction-tuning. This two-stage training scheme is inspired by curriculum learning, which progressively enhances the medical proficiency of VL models.
\subsubsection{Feature Alignment}

In this first stage, Qilin-Med-VL is trained on an image description task, where the model is asked to predict a caption for each input medical image. For each pair of medical images and text in the dataset, we instructed the model to generate a caption for the image(prompts summarized in Appendix.~\ref{fig:prompts}. We used the actual captions as the correct answers during training. Importantly, we fix the parameters of the pre-trained image encoder and language model (LLM). Instead, we train a special adapter to make sure visual and language features representing the same medical concepts align well. This alignment helps the model better understand medical information across different forms (visual and text) and improves the consistency of its medical concept understanding.

\subsubsection{Instruction-Tuning}

In the second stage, we further improved Qilin-Med-VL's capability of following medical instructions. We used a dataset specifically designed for this purpose, as discussed in Sec.~\ref{stage2_data}. In this stage, each training example consisted of a medical image and a related question. The model's task was to answer the question using the information in the image. We freezed the pre-trained image encoder and fine-tuned the language model and the vision-language feature adapter. This way, Qilin-Med-VL becomes more proficient at understanding various medical instructions and can carry out medical tasks, like answering medical questions based on images, in a conversational manner.

\section{Experiments}



    




\subsection{Baselines}
To investigate Qilin-Med-VL's ability in medical vision-language understanding and instruction following, we conduct a comparative analysis with five baseline LMMs:
\begin{itemize}
    \item \textbf{GPT-4V}\cite{openai2023gpt4}, a large multi-modal model that, while less capable than humans in many real-world scenarios, exhibits human-level performance on various professional and academic benchmarks.
    \item \textbf{Qwen-VL} \cite{Bai2023QwenVLAF}, an open-sourced general large vision-language model based on Qwen-7B\cite{bai2023qwen} and ViT\cite{dosovitskiy2021image} that can handle various vision-language tasks, including image description, question-answering, grounding, and text-reading.
    \item \textbf{VisCPM-Chat} \cite{hu2023large}, trained using CPM-Bee\footnote{https://github.com/OpenBMB/CPM-Bee/tree/main} with 10B parameters, fusing visual encoder Muffin\cite{yu2023reformulating} and visual decoder Diffusion-UNet\cite{rombach2022highresolution} to support visual inputs and outputs.
    \item\textbf{LLaVA-1.5} \cite{liu2023visual}, an open-sourced end-to-end trained LMM based on Vicuna-13B\cite{vicuna2023} and ViT\cite{dosovitskiy2021image}.
\end{itemize}

\subsection{Implementation Details}
We used Chinese-LLaMA2-13B-Chat as the foundation LLM and Clip-ViT-large-patch14-336 as the pre-trained image encoder for Qilin-Med-VL. Chinese-LLaMA2-13B-Chat is an open-source Transformer-based LLM with 13 billion parameters further trained on Chinese-LLaMA2-13B, which is optimized for conversation scenarios. Clip-ViT-large-patch14-336 is a pre-trained CLIP vision encoder trained by OpenAI. 

We performed two-stage curriculum training using 8 $\times$ A100 80G GPUs. Specifically, We had the following settings during feature alignment: batch size = 32 per GPU, 1 epoch, learning rate = 1e-3, warmup ratio = 0.03, and max length = 2048. As for the vision-language instruction-tuning stage, we used the following settings: batch size = 16 per GPU, 1 epoch, learning rate =  2e-5, warmup ratio = 0.03, and max length = 2048.

\begin{figure*}
\begin{minipage}{1.0\textwidth}
  \strut\newline
  \centering
  \includegraphics[width=0.9\textwidth]{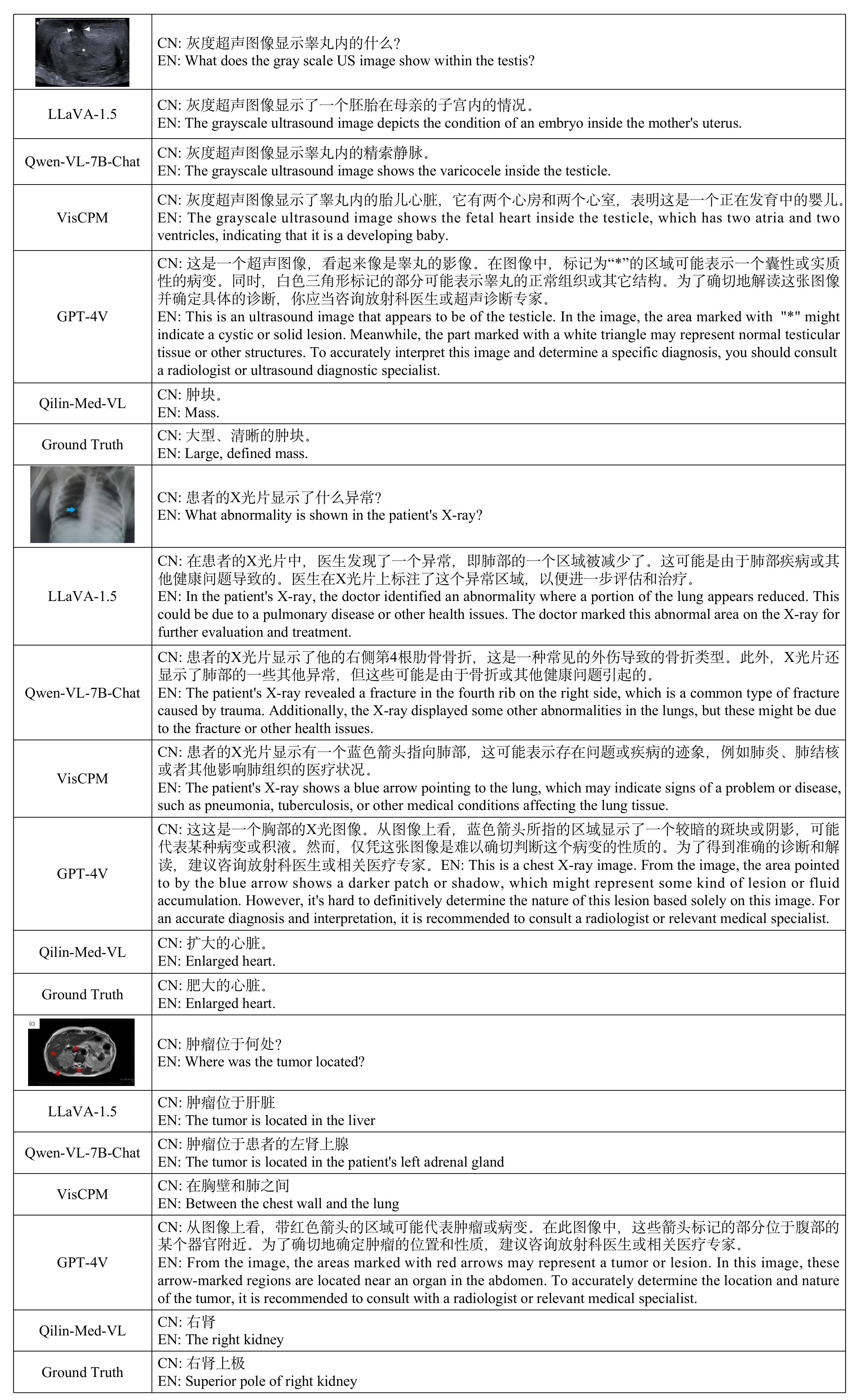}
  \captionof{figure}{Case Study of Qilin-Med-VL and baselines.}\label{fig:results_2}
\end{minipage}
\end{figure*}


\subsection{Results and Discussion}

Fig.\ref{fig:results_2} shows some results of Qilin-Med-VL and various baselines on the PMC-VQA test set. We display cases of different types of images, including ultrasound, X-ray, MRI, etc. 

For the first case, even though the image is clearly informed to be related to the testis, LLaVA still determined it to be an embryo in the uterus. Qwen-VL predicted it to be a varicocele inside the testicle. VisCPM made a fundamental mistake, predicting that there was a fetus inside the testicle and describing the specific situation. GPT-4V's answer was relatively accurate, suggesting the possibility of a cystic or solid lesion. In contrast, Qilin-Med-VL accurately predicted that there was a tumor in the region. 


For the second case, both LLaVa and VisCPM suggested abnormalities in the lungs, while Qwen-VL suggested there was a rib fracture. GPT-4V did not give a clear judgment. However, Qilin-Med-VL predicted the abnormality to be an enlarged heart. 

For the third evaluative task, we provided the models with clinical information indicating the presence of a pathological condition and challenged them to ascertain the tumor's anatomical location based on the imaging data. LLaVA, Qwen-VL, and VisCPM misidentified the site of the lesion. GPT rendered a non-specific interpretation, suggesting the tumor's presence within an organ in the abdominal region, yet without precise localization. Conversely, Qilin-Med-VL demonstrated precision by accurately pinpointing the tumor to the right renal region. 

We sought the expertise of a medical specialist who conducted a meticulous analysis based on the image data. The specialist astutely observed that the liver was located in the upper left quadrant of the image, while the kidneys were bilaterally aligned adjacent to the spinal column. This comprehensive evaluation, considering both the tumor's position and morphology, led the specialist to the conclusion that the tumor was localized within the renal region.









\section{Limitations}
We acknowledge that this study, as the pioneering effort in deploying large vision-language models in the Chinese healthcare sector, has a few limitations that need to be addressed in future research. A critical limitation is the study's dependence on machine-translated data., which could inadvertently introduce biases or inaccuracies, affecting the model's reliability. This limitation also underscores the importance of linguistic and cultural sensitivity in healthcare applications and the need for rigorous validation methods to ensure the accuracy of generated and translated content. Additionally, the absence of multi-turn conversation data in the current dataset limits the model's ability to handle complex, multi-round interactions effectively. 

\section{Ethics and Societal Impacts}
The development and deployment of LLMs and large vision-language models in healthcare present various ethical considerations and potential societal impacts. A primary concern is these models lack comprehensive clinical validation and are only for academic and research purposes. As such, Qilin-Med-VL should not be employed for medical advice or healthcare decisions at this stage, as misuse could lead to incorrect or harmful outcomes.

In navigating the intersection of artificial intelligence and healthcare, upholding ethical principles, prioritizing patient safety, data privacy, and equitable technology access is essential. Qilin-Med-VL represents a promising advancement but is just one step toward universally accessible healthcare AI solutions. Its ethical responsibility and clinical validation for real-world applications remain to be demonstrated.

\section{Conclusion \& Future Work}
The development of Qilin-Med-VL represents a pioneering step in integrating advanced large vision-language models for Chinese healthcare. This research underscores the importance of linguistic inclusion and the need for specialized models in non-English-speaking communities. We anticipate future research to continue to refine this field, with the ultimate goal of democratizing healthcare access and elevating global health standards with the help of medical AI.

\bibliography{custom}

\appendix
\clearpage

\section{Appendix}
\subsection{Prompts for medical feature alignment and instruction tuning}
\begin{minipage}{1.0\textwidth}
  \strut\newline
  \centering
\includegraphics[width=1.0\textwidth]{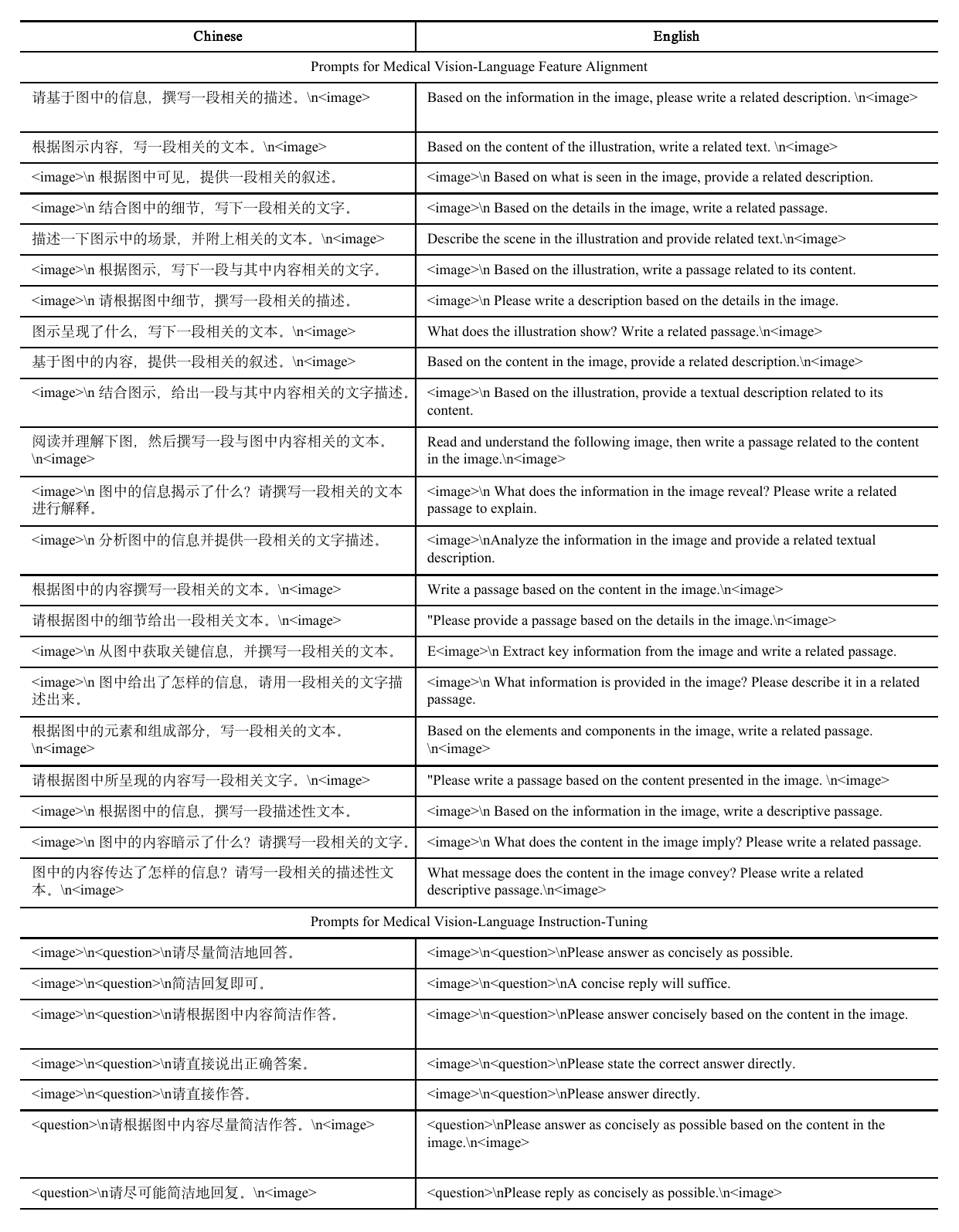}
  \captionof{figure}{Prompts for medical feature alignment and instruction tuning.}\label{fig:prompts}
\end{minipage}



\end{document}